\documentclass[11pt]{article}
\usepackage{geometry}
\usepackage{coling2020}
\usepackage{times}
\usepackage{url}
\usepackage{latexsym}
\usepackage{microtype}
\usepackage{graphicx}
\usepackage{amsmath}
\usepackage{amssymb}

\colingfinalcopy 

\title{KdeHumor at SemEval-2020 Task 7: A Neural Network Model for
Detecting Funniness in Dataset Humicroedit}

\author{Rida Miraj  \and Masaki Aono  \\
  Department of Computer Science and Engineering \\
  Toyohashi University of Technology, Japan. \\ 
  {\tt ridamiraj974@gmail.com} ,
  {\tt aono@tut.jp} \\}

\date{}

\begin{document}
\maketitle
\begin{abstract}
  This paper describes our contribution to SemEval-2020 Task 7:  Assessing Humor in Edited News Headlines. Here we present a method based on a deep neural network. In recent years, quite some attention has been devoted to humor production and perception. Our team KdeHumor employs recurrent neural network models including Bi-Directional LSTMs (BiLSTMs). Moreover, we utilize the state-of-the-art pre-trained sentence embedding techniques. We analyze the performance of our method and demonstrate the contribution of each component of our architecture.
\end{abstract}

\blfootnote{
    %
    %
    \hspace{-0.65cm}  
    This work is licensed under a Creative Commons Attribution 4.0 International Licence.
    Licence details:
    \url{http://creativecommons.org/licenses/by/4.0/}.
   
}

\section{Introduction}
\label{sec:introduction}
Humor, a way of giving entertainment and provoking laughter, is an exceptionally savvy communicative activity. In recent years, quite some attention has been devoted to humor production and perception. What makes us laugh when reading a funny sentence?  Mostly it depends on the cultural area background of a person. Humor cannot be defined in standards as it varies with nation to nation, area to area, people to people. Many types of humor require substantial external knowledge such as irony, wordplay, metaphor and sarcasm. These factors make the task of humor recognition difficult.

Recently, with the advance of deep learning that allows end-to-end training with big data without human intervention of feature selection, humor recognition becomes promising. However detecting humor in news headlines from which one word is chosen to be replaced by another word to make the micro-edited sentences look funny is quite challenging. To address the challenges of humor detection in edited news headlines, ~\cite{SemEval2020Task7} proposed  detection of funniness level  in dataset named as Humicroedit\footnote{Dataset formed for this task is named as humicroedit in the article ~\cite{hossain-etal-2019-president}.}, Task 7 at SemEval-2020.
They focus on two related subtasks. Task 1 defines regression problem where a system needs to predict how much funny is the edited news headlines (humicroedit) from range 0 to 3. Whereas Task 2 defines the classification problem where a system need to predict the funniest sentence between two edited headlines. 
In this paper, we only focus on the Humicroedit data for both Tasks 1 and 2. No extra data, or other sources is used to get more data. The rest of the paper is structured as follows: \textbf{Section~\ref{sec:related work}} provides a brief overview of prior research. In \textbf{Section~\ref{sec:proposed model}}, we introduce our proposed neural network model. \textbf{Section~\ref{sec:ExperimentsEvaluations}} includes experiments and evaluations as well as the analysis of our proposed method. Some concluded remarks and future directions of our work are described in \textbf{Section~\ref{sec:Conclusion}}.

\section{Related Research}
\label{sec:related work}

%
%
\blfootnote{
    %
    %
    \hspace{-0.65cm}  

    %
    %
    %
    %
}

Humor is omnipresent, universal, elusive event which exists in all societies and cultures and fulfills a range of social, cognitive and emotional functions. The task of automatic humor recognition refers to deciding whether a given sentence expresses a certain degree of humor. In previous researches and studies, mostly the problems related to humor were based on binary classification, or based on selection on linguistic features. Taylor and Mazlack analyzed on a specific type of humor. The methodology which they used was  based on the extraction of structural patterns and peculiar structure of jokes newcite~\cite{Taylor2000}. 
Purandare and Litman used humorous spoken conversations as data from a classic comedy television. To identify humorous speech in the conversation, they used standard supervised learning classifiers~\cite{purandare2006humor}.
Besides, a work has been done to discover latent semantic structure behind humor from four perspectives: incongruity, ambiguity, interpersonal effect and phonetic style. For that Yang created a computational model. He also formulated a classifier to distinguish between humorous and non-humorous instances~\cite{Yang2015}.

In other researches, with the development of artificial neural networks, many studies utilize the methods for humor recognition. Luke de Oliveira and Alfredo applied recurrent neural network (RNN) to humor detection from reviews in Yelp dataset. They also applied convolutional neural networks (CNNs) to train a model and the results from model trained with CNNs have more accurate humor recognition~\cite{de2015humor}.
Recently, Bertero and Fung proposed a first-ever attempt to employ a Long ShortTerm memory(LSTM) based framework to predict humor in dialogues. They showed how the LSTM effectively models the setup-punchline relation reducing the number of false positives and increasing the recall~\cite{Bertero2016a}.
In a recent work, Chen and Lee predicted audience’s laughter also using convolutional neural network. Their work gets higher detection accuracy and is able to learn essential feature automatically~\cite{Chen2018a}.
 Among several prominent works,  Peng-Yu Chen and Von-Wun Soo build the humor recognizer by using CNNs with extensive filter size and number, and the result shows higher accuracy from previous CNNs models by conducting experiments on two different datasets.~\cite{Chen2018}.

\section{Proposed Model}
\label{sec:proposed model}
In this section, we describe the details of our proposed neural network model. The goal of our proposed approach is to predict how much funny the edited news headlines (humicroedit) is. We tried to train our model after data processing. Figure 1 depicts an overview of our proposed model.

\begin{figure}[h!]
\begin{center}
\centerline{\includegraphics[width=1\linewidth]{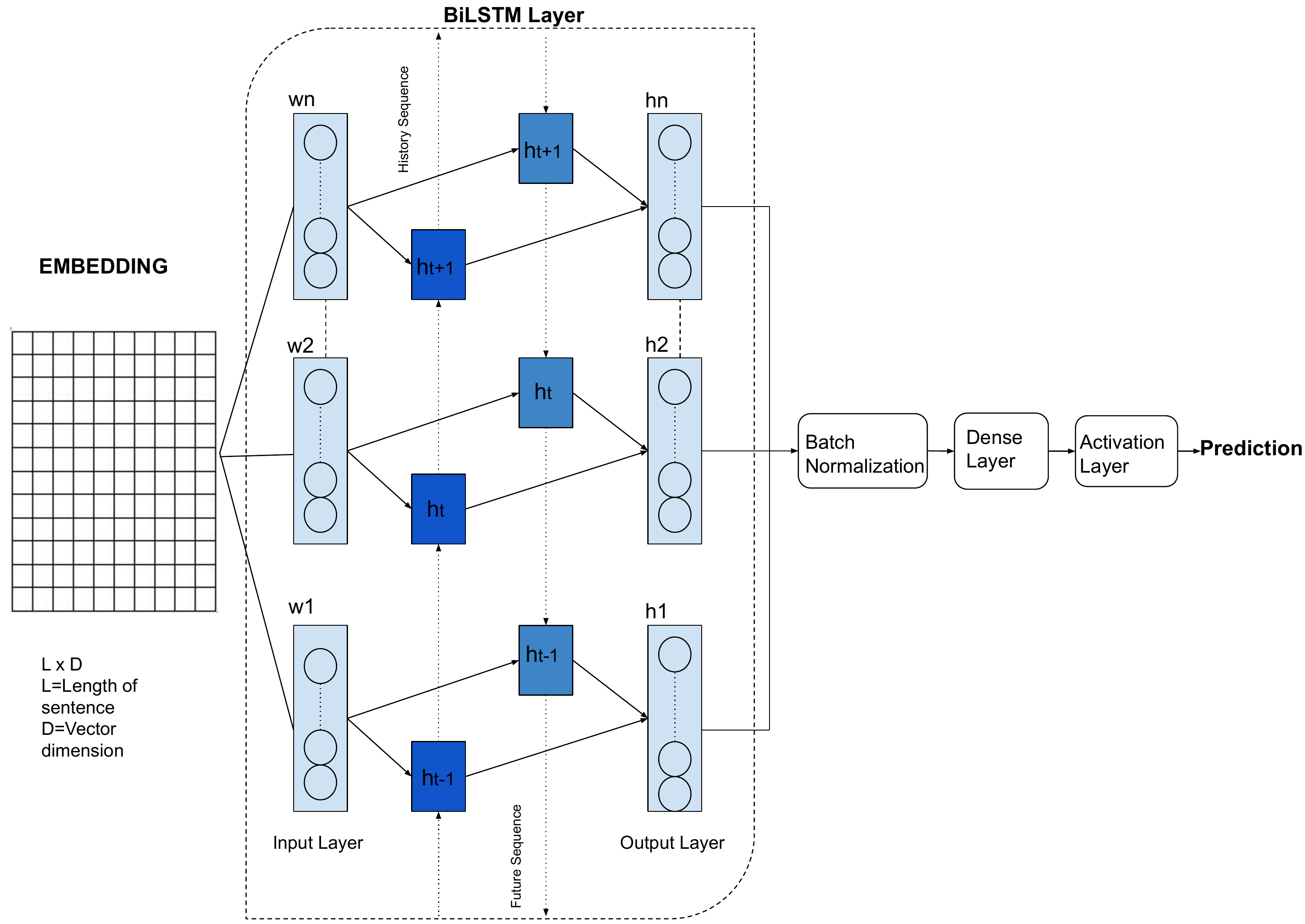}}
\caption{Proposed Model} \label{Figure 1}
\end{center}
\end{figure}

At first, in a dataset, original sentences were changes by using edit words. Then on edited sentences, pre-trained word embedding model was implemented to obtain the high-quality distributed vector representations for our datasets. Next, we apply the Bidirectional LSTMs (BiLSTMs) models to extract the higher level feature sequences with sequential information from the edited news headlines embedding. We employ an encoder pre-trained by Google News word2vec model to encode each word into 300-dimensional feature vector. These feature vectors are then sent to a Bidirectional LSTM module. Finally, the generated output feature sequences from Bidirectional LSTMs fed into the fully-connected prediction module to determine the prediction. Next, we describe each component elaborately.

\subsection{Embedding Layer}
\label{ref:embeddingLayer}
The Embedding layer in the leftmost part of Figure1 is initialized with random weights and will learn an embedding for all of the words in the training dataset. This flexible layer is defined as the first hidden layer of a network.
A pre-trained model used in embedding layer is nothing more than a file containing tokens and their associated word vectors. The pre-trained Google word2vec model was trained on Google news data (about 100 billion words); it contains 3 million words and phrases and was fit using 300-dimensional word vectors with the size of 1.53 GB.

In our proposed framework, we utilize a pre-trained word embedding model based on google news to obtain the high-quality distributed vector representations of headlines. The dimensionality of the
Embedding matrix will be $L \times D$, where $L$ is the sentence length, and $D$ is the word-vector dimension.

\subsection{Bidirectional LSTMs}
\label{ref:BiLstmLayer}
Bidirectional Long Short Term Memory (BiLSTM) in the central part of Figure1 is a bidirectional version of LSTM. BiLSTM combines the forward hidden layer with the backward hidden layer, which can access both the preceding and succeeding contexts.
BiLSTM neural network is resorted to attain the vector representation of the input sentence, which can effectively capture the semantics of the sentence.
The final  result from output layer of BiLSTM is generated through combining the results produced by both RNN hidden layers i.e. forward layer and backward layer.  Following equations show the mathematical formulation behind setting up the bidirectional hidden layer, where $V$ is for vocabulary, $b$ as bias term of weight matrix $W$. The only difference between these two relationships is in the direction of recursing through the corpus. 
$$\overrightarrow{h} _t=f(\overrightarrow{W} x_t+\overrightarrow{V}\overrightarrow{h}_{t-1}+\overrightarrow{b})$$
 \label{bilstmeq}
$$\overleftarrow{h} _t=f(\overleftarrow{W} x_t+\overleftarrow{V}\overleftarrow{h}_{t-1}+\overleftarrow{b})$$
 \label{bilstmeqq}
 
\subsection{Batch Normalization Layer}
For training very deep neural networks, a technique named as Batch normalization is used that standardizes the inputs to a layer for each mini-batch. The effect of this method is to stabilize the learning process and 
to dramatically reduce the number of training epochs required to train deep networks.
The batch normalization method is used to accelerate the training of deep learning neural networks~\cite{ioffe2015batch}.
Batchnorm for short, is proposed as a technique to help coordinate the update of multiple layers in the model. It provides an elegant way to redesign or reparametrizing almost every deep network. The reparametrization eliminates the issue of organizing updates over several layers.

Essentially, when applying this layer with BiLSTM layer, the non-linear activations of the LSTM are removed but not the gate activations. To increase the stability of a neural network, batch normalization normalizes the output of a previous activation layer by subtracting the batch mean and dividing by the batch standard deviation.

\subsection{Prediction module}
In the end a linear activation layer is used as our task is a regression type problem. The final layer of the neural network will have one neuron and the value it returns is a continuous numerical value. Using RELU function as activation layer can be a good option as the output of our task is non-negative real numbers. For loss function, we employ mean square error (MSE), which is defined as follows: 
$$ MSE = \frac{1}{n} \sum_{i=1}^{n} (y_i - \hat{y}_i)^2 $$, 
where
$n$ is the number of samples, $y_{i}$ is the true value of $i$-th data, and $\hat{y_{i}}$ is the predicted value of $i$-th data.
For optimization, we  adopt RMSprop (Room Mean Square propagation), as RMSprop is adaptive and fast. In RMSprop, instead of taking cumulative sum of squared gradients it takes the exponential moving average of gradients \cite{Medium}.

\section{Experiments and Evaluations}
\label{sec:ExperimentsEvaluations}
\subsection{Dataset Collection}
\label{ref:datasetCollection}
For Sub-task-1, two training datasets named Humicroedit ~\cite{hossain-etal-2019-president}  and Funlines ~\cite{hossain-etal-2020-stimulating}  were provided. But only Humicroedits data was used. The training, validation, and test set of the news headline dataset contains the 9653, 2420 and 2025 humicroedits, respectively.
\subsection{Model configuration}
\label{ref:modelConfiguration}
In the following, we describe the set of parameters
that we have used to design our proposed
neural network model. The framework which we used to designed our model was based on TensorFlow~\cite{abadi2016tensorflow} and training of our model is done on a GPU~\cite{owens2008gpu} to capture the benefit from the efficiency
of parallel computation of tensors. 
We used the 300-dimensional word2vec Google news pre-trained model, where the maximum length of sentence in given data was 20.
We trained all models with 100 epochs with a batch size of 128 and an initial
learning rate 0.001 by RMSprop optimizer. Unless
otherwise stated, default settings were used for the
other parameters.

\subsection{Evaluation Results}
\label{ref:evaluationMeasure}
To evaluate the performance of the system, the organizers
used different strategies and metrics for
the Sub-tasks 1 and 2~\cite{SemEval2020Task7}. For the Sub-task
1, Root Mean Square (RMSE), RMSE@10, RMSE@20 and RMSE@30 were applied
to estimate the performance of a system. Among three evaluation measures, RMSE was considered as the primary evaluation
measure for this task.

\begin{figure}[h!]
\begin{center}
\centerline{\includegraphics[width=0.6\linewidth]{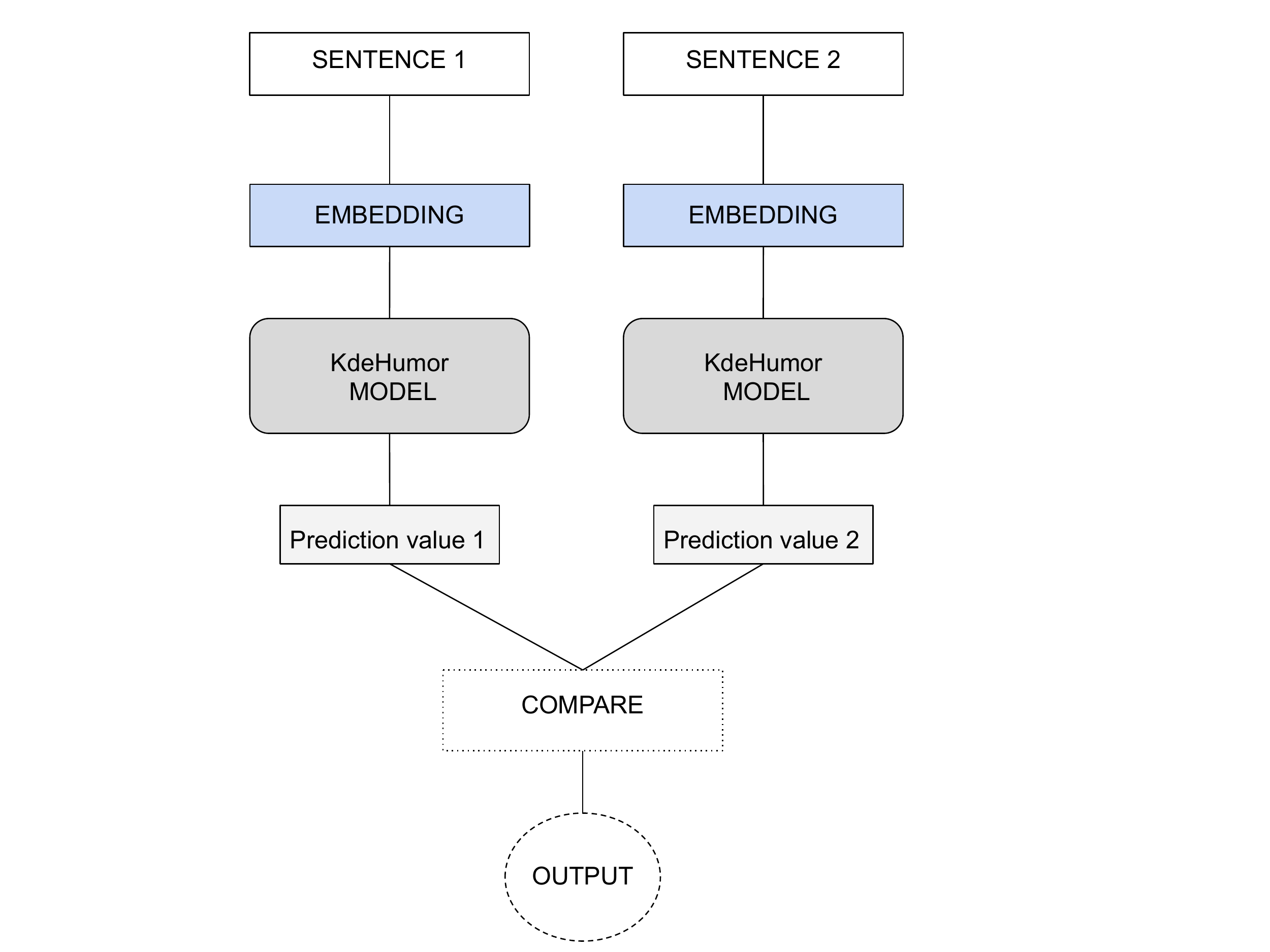}}
\caption{\small Our approach for Sub-task 2.} \label{Figure 2}
\end{center}
\end{figure}
Figure 2 illustrates our approach used for Sub-task 2. In our approach, given two edited headlines, we apply KdeHumor Model from Sub-task 1 to each edited headline, and decides which edited headline is funnier by comparing the predicted values. In  Sub-task 2,  model Accuracy is used as the evaluation measure. Another auxiliary metric called the Reward is used. For a larger funniness difference between the two headlines in a pair, Reward is higher for a correct classification.

\subsection{Experimental Results }
We now evaluate the performance of our proposed
method in this section.
The summarized results for Sub-task 1 and 2
are presented in Tables 1 and 2, respectively.
\begin{table}[h]
\begin{scriptsize}
\begin{center}
\begin{tabular}{lrl}
\hline \bf Team Name & \bf RMSE & \bf RMSE@10 \\ 
\hline KdeHumor & 0.6164 & 1.0175 \\ \hline
Farah & 0.5339 &0.9091  \\
HonoMi & 0.4972 & 0.7791 \\
Heidy & 0.6833 & 1.0496 \\
frietz58& 0.7225 & 1.0634 \\
zxchen & 0.5288  & 0.8636 \\
Mjason & 0.5351 & 0.8748 \\

\hline
\end{tabular}
\end{center}
\caption{\label{task-1} (Sub-task 1) KdeHumor results with other selected team. }
\end{scriptsize}
\end{table}
At first, we presented the performance of our
proposed method denoted by team name KdeHumor as well as presenting the performance of randomly
chosen top-ranked participated systems.

\begin{table}[h]
\begin{scriptsize}
\begin{center}
\begin{tabular}{lrl}
\hline \bf Team Name & \bf Accuracy & \bf Reward \\ 
\hline KdeHumor &0.5190 & 0.0271 \\ \hline
Farah & 0.6088 &0.1840  \\
HonoMi & 0.6742 & 0.2987 \\
HumorAAC & 0.3204 & -0.2177 \\
Heidy& 0.4197 & -.0990 \\
\hline
\end{tabular}
\end{center}
\caption{\label{task-2} (Sub-task 2) KdeHumor results with other selected team. }
\end{scriptsize}
\end{table}

\section{Conclusion}
\label{sec:Conclusion}
In this paper, we presented our approach for 
SemEval-2020 Task 7: Assessing Humor in Edited News Headlines. Detection of humor is not an easy task. We tried to tackle the problem by employing deep learning techniques. 
We conducted some experiments using different models for example we implemented simple regression model, multilayer perceptron model but the model that gave better result was KdeHumor model.
In future, we are planning to incorporate BERT model for further improvement.

\section*{Acknowledgements}
This work was partially supported by JSPS KAKENHI Grant Number 17H01746.

\bibliographystyle{semeval2020}
\bibliography{semeval2020}

\end{document}